\documentclass[runningheads]{llncs}
\usepackage{float}
\floatstyle{plaintop}
\restylefloat{table}
\usepackage{graphicx}
\usepackage{comment}
\usepackage{amsmath,amssymb,commath}
\usepackage{color}
\usepackage{subfig}
\usepackage{url}
\usepackage{hyperref}

\newif\ifreview
\reviewfalse

\ifreview
	\usepackage{lineno}

	\linenumbers
\fi

\begin{document}

%%%%%%%%%%%%%%%%%%%%% Add submission id, track, and title. %%%%%%%%%%%%%%%%%%%%%

% Insert your submission number here
\def\SubNumber{ 83 }

% Choose one track by uncommenting one of the following lines  
\def\GCPRTrack{83}
% \def\GCPRTrack{Track: Computer vision systems and applications}
% \def\GCPRTrack{Track: Pattern recognition in the life and natural sciences}
% \def\GCPRTrack{Track: Photogrammetry and remote sensing}
% \def\GCPRTrack{Track: Robot vision}

% Replace with your title
\title{FasterVideo:  Efficient Online Joint Object Detection And Tracking}
% You can use \thanks for acknowledgment. Do not add any acknowledgment to the draft 
% version that is used for the review process.  
%\title{Title\thanks{XXX}}

\ifreview
	% ANONYMOUS SUBMISSION FOR REVIEW
	% DO NOT MODIFY these for the draft version that is used for the review process.
	\titlerunning{}
	\authorrunning{BLIND Submission \SubNumber{}. CONFIDENTIAL REVIEW COPY.}
	\author{BLIND SUBMISSION - \GCPRTrack{}}
	\institute{\SubNumber}
\else

	\author{Issa Mouawad\inst{* 1,2}\orcidID{0000-0002-5682-4733}\and
	Francesca Odone \inst{1,2}\orcidID{0000-0002-3463-2263} }
	
	\authorrunning{I. Mouawad et al.}
	
	\institute{MaLGa Machine Learning Genoa Center \and 
	DIBRIS - Università degli Studi di Genova 
	\email{issa.mouawad@dibris.unige.it,francesca.odone@unige.it}\\
%	\url{http://www.springer.com/gp/computer-science/lncs} \and ABC Institute, Rupert-Karls-University Heidelberg, Heidelberg, Germany\\
%	\email{\{abc,lncs\}@uni-heidelberg.de}
}
\fi

\maketitle              % typeset the header of the contribution

\begin{abstract}
Object detection and tracking in videos represent  essential and computationally demanding building blocks for current and future 
visual perception systems.  In order to reduce the efficiency gap between available methods and computational requirements of real-world applications, we propose to re-think one of the most successful methods for image object detection,  Faster R-CNN, and extend it to the video domain.  Specifically, we extend the detection framework to learn instance-level embeddings which prove beneficial for data association and re-identification purposes.  
Focusing on the computational aspects of detection and tracking, our proposed method reaches a very high computational efficiency necessary for relevant applications, while still managing to compete with recent and state-of-the-art methods as shown in the experiments we conduct on standard object tracking benchmarks\footnote{Code available at {\tt https://github.com/Malga-Vision/fastervideo}}.
\end{abstract}

\section{Introduction}
 Detecting and tracking  multiple objects in video sequences is a core building block for several applications. 
 Recently, deep learning based methods achieved unprecedented success in detecting objects in both general-purpose and application-oriented settings 
 \cite{ren2015faster,yolo,centernet}. Utilizing such methods in the video domain remain challenging due to the inefficiency of per-frame processing and the lack of a temporally consistent understanding of objects trajectories.

Multiple-object tracking deals with the task of tracking several targets locations across video frames and 
is able to derive trajectories across time by associating tracks and detected objects. Multiple-object tracking, however, 
is usually handled in the tracking-by-detection framework, which assumes that objects in each frame are detected using a separate algorithm and only addresses  association \cite{gnn3dmot,dhn}. This separation causes an additional computational cost, and prohibits the sharing of representations and information between the two tasks.

Recently, it has been noted that joint architectures  reduce the computational overhead by deriving the outputs of multiple tasks simultaneously, proving in the same time the performance benefits of this holistic processing where computation and hidden representations are shared among tasks \cite{taskonomy,multinet,blitznet}.  

Building on this line of research, we address detection and tracking jointly,  relying on a state of the art image object detector, Faster R-CNN \cite{ren2015faster}, which we extend  to the video domain. 
\begin{figure*}
\center
      \includegraphics[height=5cm]{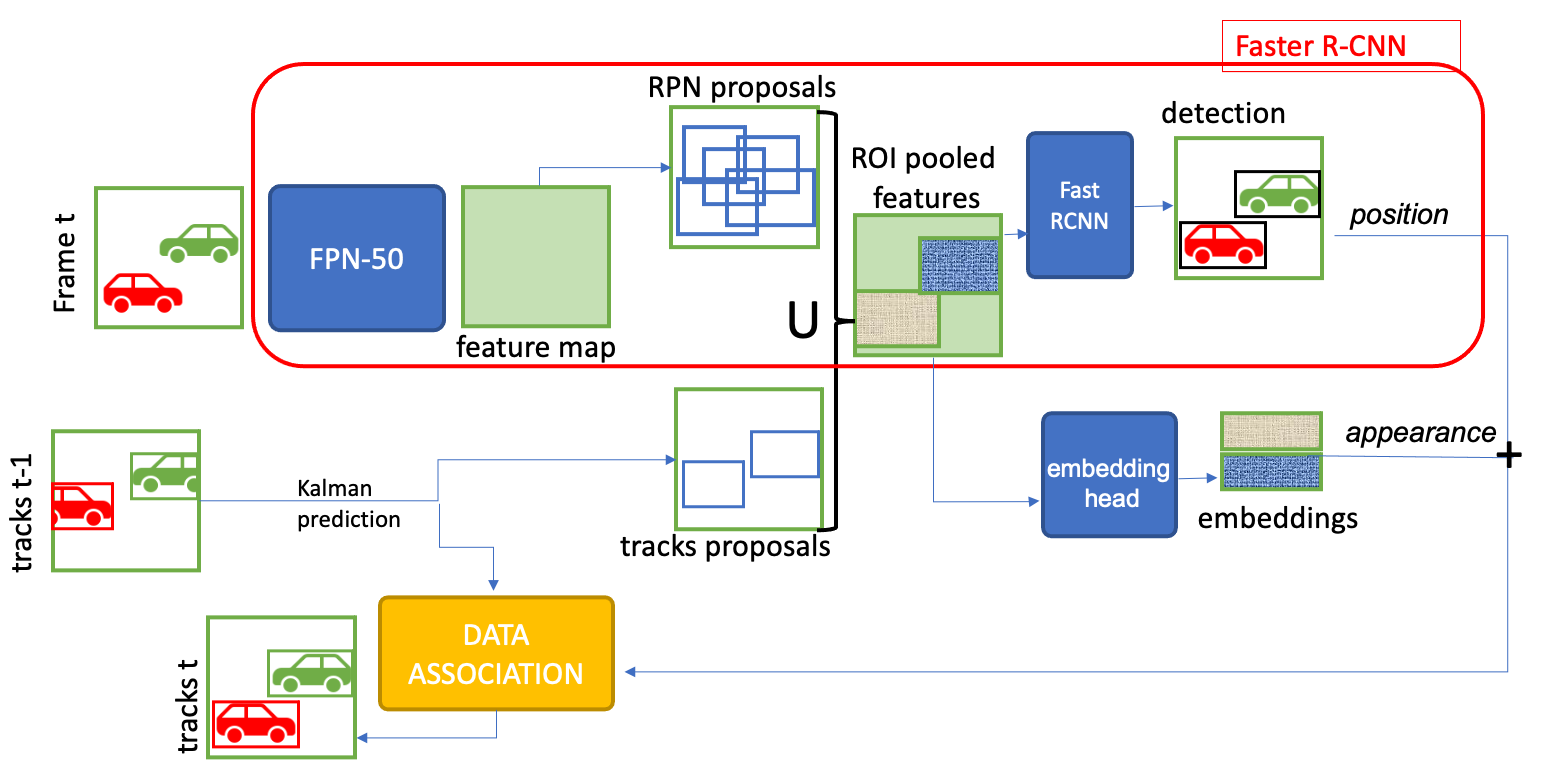}
      \caption{Overview of our proposed method for joint detection and tracking}
      \label{fig:method}
  \end{figure*}
  
% We take advantage of the state of the art in image detection, building our joint detection and tracking procedure as an extension of  Faster R-CNN \cite{faster}.
 The modular structure of two-stages detectors, allows us to control the computational cost of the detector, while exploiting additional information provided by the video input: we control the complexity of Faster R-CNN Region Proposal Network (RPN) by reducing of the number of image-based proposals, while adding proposals originated by the previous video frames. Also, the two-stage design can be extended to learn appropriate embeddings which are effective to improve the tracking association step ---see Figure \ref{fig:method}.
 
In summary, the contributions of our paper can be summarized as follows:
\begin{itemize}
    \item We provide a modular extension of a 2-stage image-based object detector to address jointly video object detection and tracking. 
    \item We re-use the detector learned representations to simplify the tracking task by learning embeddings which are then employed to boost the data association accuracy across frames. 
    \item We use temporal prior of objects location to both guide the detector and reduce its computational cost
    \item Our pipeline achieves accuracy results comparable to state-of-the-art methods, while consistently delivering several fold efficiency improvement , highlighting under-explored accuracy-speed trade-off points.
\end{itemize}

The rest of this paper is organized as follows: first, we review related works tackling detection and tracking tasks either separately or jointly. Next, we describe in details our proposed pipeline targeted at joint detection and tracking. Finally, we conduct experiments using KITTI \cite{kitti} and MOT \cite{mot,mot20} datasets and we compare with other similar methods in terms of both MOT metrics and inference time to highlight speed performance trade-offs.

\section{Related Work}

%{\color{red} [THIS SECTION MUST BE SHORTENED (AT MOST ONE PAGE MAYBE LESS. I WILL DO IT AS THE PAPER EVOLVES]}
\subsection{Video Object Detection}

Unlike image-based methods which can be applied in video settings on separate frames independently, several works try to rely on additional video prior to guide video object detection. This direction is gaining interest due to the challenges posed by videos to image-based detectors such as motion blur and focus loss.

In the work of \cite{Zhu_2018_CVPR}, optical flow is used as to warp features from consecutive frames, while a recurrent neural network is used in \cite{spatiotemporal} to aggregate frames and learn to detect objects across time. These methods, however, do not construct trajectories of detected objects.
\subsection{Multi-object Tracking}

Due to the high accuracy achieved by recent object detectors, many works in the literature follow the tracking-by-detection approach. Such approach assumes that detections are obtained for each frame separately, and focuses on association.
Association between detections and tracks is often formulated as a bipartite graph and an association cost is identified for each possible match. The problem is usually solved using the Hungarian algorithm \cite{hungarian}.
In addition to intersection over union (IoU) metric, 
 several approaches propose the use of appearance-based similarity for matching, obtained using optical flow, low-level feature descriptors, or motion-based features \cite{mass,smat,tusmiple}. While other methods try to learn similarity metrics from data using contrastive learning \cite{quasidense} or triplet losses \cite{nobells,lpc} Different data representation is additionally explored in recent works \cite{dhn,gnn3dmot,weng2020joint} allowing end-to-end learning of MOT methods.

\subsection{Joint Object Detection and Tracking}
This research direction focuses, instead, jointly on both detection and tracking tasks. 
These methods, in particular, promise to simplify perception tasks and yield more efficient pipelines.
In \cite{detectortrack}, the authors propose a scheduler network which is used to alternate between running a full detector or simply locating already-seen objects.
Another direction aims to transform typical object detectors to perform jointly detection and tracking. 
Some of these methods extend a two-stage detector (mainly Faster R-CNN) \cite{nobells,mots}, while others focus on single-stage detectors \cite{centertrack,retinatrack}. Most of these methods, however, introduce additional modules which reduce efficiency of the joint pipeline, even below that of the baseline detector. 

\section{Proposed Method}
 In this section we present our architecture based on the well established Faster R-CNN object detector, which we extend to address detection and tracking jointly and efficiently. By reducing the number of proposals and relying on  tracked objects as an additional source of proposals, we improve the speed-accuracy trade-off. Additionally, we 
 introduce an embedding network branch that allows us to learn an appropriate and descriptive appearance representation for tracked instances boosting data association accuracy. 
Figure \ref{fig:method} shows the building blocks of the proposed method.% discussed throughout the section.

\subsection{Video Object Detection}
We adopt as an image-based object detector Faster R-CNN with a FPN-50 backbone and pyramid layers \cite{fpn}.
To optimize it for the video-based object detection, we propose two modifications. 
\\

\begin{figure}
\centering
\parbox{5cm}{
      \centering

      \includegraphics[height=3.7cm]{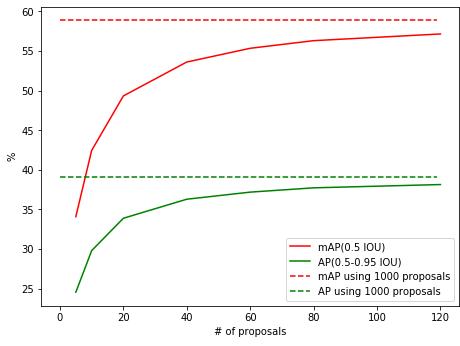}
      \caption{Accuracy obtained with different number of proposals  (COCO2017 validation set). }
      
      \label{fig:coco_acc}
  
 }
 \qquad
 \begin{minipage}{.5\textwidth}
 \centering
     \includegraphics[height=3.7cm]{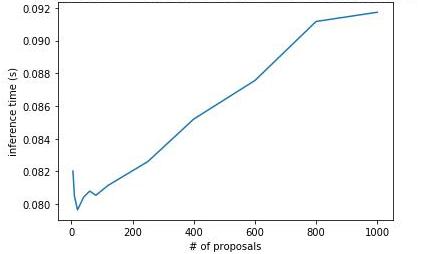}
      \caption{ Inference speed with different number of proposals reoprted on Quadro P-5000 }
      \label{fig:coco_time}
       \end{minipage}
  \end{figure}

\noindent {\bf Sparser RPN Proposals.} 
Sparser proposals allow us to control the computational cost of the detector. Figures \ref{fig:coco_acc} and \ref{fig:coco_time} provide an experimental evidence:
 %We use a reduced number of proposals generated by the RPN compared to the original implementation to control the computation overhead associated with running these proposals down the Fast R-CNN head. 
%The relation between the number of proposals and both the inference speed and the detection accuracy has been studied thoroughly while taking this decision. 
%
 Figure \ref{fig:coco_acc},  reports the accuracy (measured in both mAP and AP) on COCO2017 validation set using different number of proposals compared with the original 1k proposals \cite{ren2015faster}. Comparable accuracies are obtained with $1/10$ of the proposals.  Figure \ref{fig:coco_time} shows a slow yet consistent decrease in the inference time  incurred by decreasing the number of proposals.
%These experiments suggest that using few proposals, we may be able to recover a high accuracy while cutting significantly the inference time.
%
To maintain a high accuracy, we  exploit space-time continuity of objects instances in subsequent frames, as we detail in the following.
 
\noindent {\bf Tracks Proposals.}
In order to compensate for the reduced number of proposals, and address occlusions and illumination changes, we provide a reverse feedback from the tracker to the detector. We perform a Kalman prediction step on tracks from the frame $t-1$ (with two opposing bounding box corners in the Kalman state) and add these predicted bounding boxes as proposals for the frame at time $t$. This provides an attention mechanism for the detector to focus on important parts of the image where previously seen objects are expected to be found. %More concretely, using tracks as proposals can help the detector in recovering heavily occluded objects in the few-proposals regime 

\subsection{Embeddings Learning}
Deep features associated with the entire image or a specific instance provide a robust descriptor which withstands a certain viewpoint or illumination change \cite{instancefeatures}. 
For this, we add an embedding learning branch to the classical Faster R-CNN RoI Heads, extracting a representation for any object instance which we will employ for data association and re-identification.
 Re-identification plays an important role in object tracking systems \cite{reid} but it is typically addressed as a separate task \cite{nobells}. We, instead, integrate the embedding module within our joint detection and tracking framework. This allows the network to robustly handle occlusions and fuzzy associations relying on objects encoding learned specifically to be viewpoint and illumination invariant. 

The new embedding learning branch is placed on top of the RoI-pooled features using two fully connected layers separated by a ReLU non-linearity and batch normalization.

The resulting network is trained jointly for both detection and embedding learning tasks and final loss is composed of three losses: the original RPN and Fast R-CNN losses \cite{ren2015faster}, plus the embedding loss for which we use the triplet loss \cite{facenet}. 
While RPN and detection losses are calculated as usual for all the ground truth objects present in the images, in the case of the triplet loss we need to select {meaningful triplets} to keep convergence under control \cite{facenet}. 

\begin{figure}
\center
      \includegraphics[height=3cm]{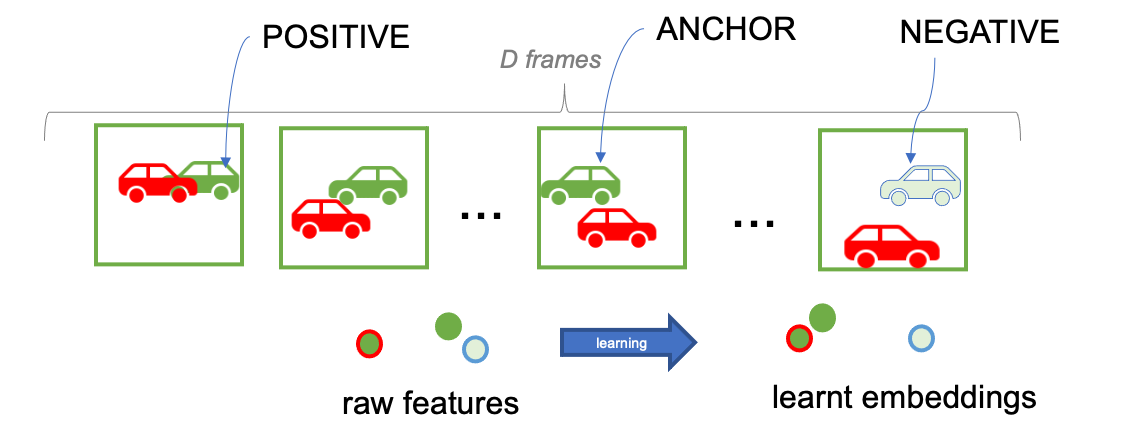}
      
      \caption{The distance of the learned embeddings is smaller for instances of the same object (anchor and positive) and larger for different objects (anchor and negative)}
      \label{fig:triplet}
  \end{figure}

An instance $x^a_i$ of a specific  object $i$ (an anchor), should be closer to the positive example $x^p_i$ (another instance of the same object) than to a negative one $x^n_i$ (an instance of another object), by a margin $\alpha$:
\begin{eqnarray}\label{eqn:1}
\norm{E(x^a_i)-E(x^p_i)}_2^2 + \alpha \textless \norm{E(x^a_i)-E(x^n_i)}_2^2.  
\end{eqnarray}
Then the triplet loss seeks to maximize the distance between the encoding of the anchor example $x_i^a$ and the negative example $x_i^n$, while, at the same time, minimizing the distance between the anchor example and the positive example $x_i^p$, as depicted in Figure \ref{fig:triplet}. 
%The difference between the mentioned distances is further emphasised by a margin $\alpha$ (which is set to 0.2 following \cite{facenet}).

The choice of the triplets is crucial as it is infeasible to consider all combinations of the whole  training set.
%, and applying it to every combination within the batch slows convergence especially for triplets which already satisfy Inequality (\ref{eqn:1}). 
We apply a batch-wise hard example mining. 
First, we choose a batch of $B=8$ images at random from a consecutive $D=16$ frames, considering only batches where at least $P=8$ objects are present at least $K=4$ times. This  allows for a robust calculation of the triplet loss; the intuition behind sampling the batch at random from neighborhood of frames is to mimic missed detections and increase the robustness of the learned embeddings against severe appearance changes.
Then, for each anchor example, we pick the hardest positive and negative examples within the batch $B$ such that:
\begin{eqnarray}
x^p = \arg \max_{x^p_i \in B}{\norm{E(x^a_i)-E(x^p_i)}_2^2} \ \ \ \mbox{and} \ \ 
x^n = \arg \min_{x^n_i \in B}{\norm{E(x^a_i)-E(x^n_i)}_2^2}
\end{eqnarray}
Finally, the loss is formulated using triplets $(x_i^a, x_i^p, x_i^n)$ for all the $P$ objects:
\begin{eqnarray}\label{eqn:loss}
L_{embedding} =\frac{1}{n} \sum_{i =1}^{P} [ \norm{E(x^a_i)-E(x^p_i)}_2^2 -\norm{E(x^a_i)-E(x^n_i)}_2^2 + \alpha]_+
\end{eqnarray}

\subsection{Data Association and Tracking}

\label{new_objects}
During inference, RoI-pooling is performed as the original Faster R-CNN using the proposals generated by RPN to extract instance-level features which are fed to the two heads (the Fast R-CNN one and the proposed embedding head depicted in Figure \ref{fig:method}). Usual class and box predictions are calculated as in \cite{fast}, while the final boxes are used to pool features which are fed to the embedding head. 
In order to assign identifiers to detected object, we formulate a data association step which matches the detections at time $t$ with the tracklets from $t-1$. We define, inspired by \cite{usv}, a matching cost between detections and tracklets as the linear combination of two distances: the Jaccard or IoU Distance to capture {\em position or spatial proximity} between bounding boxes and the cosine distance (defined as $1- $cosine similarity) between the two objects embeddings which captures {\em appearance similarity}: 
\begin{eqnarray}\label{eqn:dist}
Cost = \alpha * dist_{position} + \beta * dist_{appearance} 
\end{eqnarray}
($\alpha$ and $\beta$ are weighting factors, we both set to $0.5$).
Next,  we rely on the Hungarian algorithm \cite{hungarian} to perform a minimal-cost matching between the detector output and the tracks leading to matches and possibly mismatches.
In order to avoid forced one-to-one weak matches, we set a maximum cost for any match which we discard if violated.

\begin{figure}
\center
      \includegraphics[height=5.81cm]{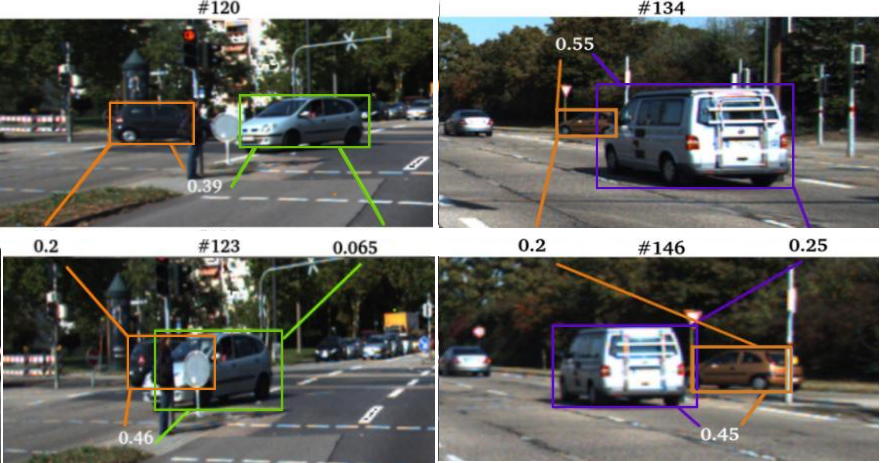}
      
      \caption{Examples of occluded object re-identification using object embeddings  and cosine distance (best seen in color), \# indicate frame number}
      \label{fig:reid}
  \end{figure}

In Figure \ref{fig:reid}, we show some examples of re-identification events accomplished utilizing objects embeddings. 
We report on the figure the embedding distance between pair of objects across different frames showing the ability of the embedding head to generate similar embeddings for the same object even across a wide time frame and for challenging appearance shifts and occlusions.  

\section{Experimental Analysis}

\subsection{KITTI Benchmark}

{\em Object Detector Initial Training:}
We focus in our experiments on car objects for both detection and tracking.
We use Faster R-CNN official implementation Detectron2 \cite{detectron2} with weights of FPN trained on COCOtrain2017. Next, we fine-tune the model on KITTI's object detection benchmark focusing only on the car class. After discarding images common to the tracking benchmark, we are left with ~4k images we split to 3k images for training the object detector for 5 epochs and 1k images for validation. We keep the original number of proposals during training, but reduce it during evaluation to 20 proposals on KITTI.

\noindent  {\em Joint Detection and Tracking: } \label{results}
%In this section, we discuss in details the experiments using the re-identification head we introduced before. 
We attach the embedding head and we finetune the pipeline using a multi-task loss (formed by the original detection loss \cite{fast}, and the embedding loss from Equation (\ref{eqn:loss}))  on a split of the tracking training set (as defined in \cite{split}) using a batch size of 8 images for 5k iterations. 

\noindent {\em Oracle Detections:}
We conduct a series of experiments to highlight the performance of the tracking branch and the embedding head. To this end, we use ground truth boxes as detections source (oracle detections) and we compare the resulting tracking results using different combinations of matching costs.
Table \ref{table:or_det} shows that the association based on position provides superior performance with respect to the learnt appearance. Combining the two, however, boosts this performance further which indicates that the learned embeddings help simplify the association even with accurate boxes. 
\begin{table}[h!]
\centering
\small
\begin{tabular}{p{4.8cm}p{1.5cm}p{1.5cm}p{1.2cm}p{1.2cm}p{1.2cm}p{1.6cm}}
 \hline
  Matching method & MOTA$\uparrow$ & MOTP$\uparrow$ & FP$\downarrow$ & FN$\downarrow$  & IDs$\downarrow$ \\ 
 \hline\hline
Position (IoU distance)  & 96.0   & 90.5   & 169   & 71 & 26   \\
\hline
Appearance (embedding distance) & 94.9  & 90.8  & 214   &88  & 38 \\
\hline
Both & {\bf 97.2}  & 90.8 & {\bf 151} & {\bf 31}  & {\bf 2}     \\
\hline
\end{tabular}
\caption{ Data association cost assessment - KITTI Benchmark on a held-out set using Oracle detections  (reporting Multiple Object Tracking Accuracy - MOTA; MOT Precision - MOTP; False Positives - FP; False Negatives - FN; IDentity switches - ID)}
\label{table:or_det}
\end{table}

\noindent {\em Ablation Study:}
We use the validation set to tune hyper-parameters and study the pipeline ablations measured by the tracking accuracy and inference time\footnote{For Detection, an NVIDIA Quadro P5000 GPU has been used to obtain the time measurements} and report the results in
 Table \ref{table:comp}. The table highlights the effect of each design choice on both FPS (frames-per-second FPS)and IDs (identity switches), where using specifically-learned features and temporal prior help to recover many identity switches with the best trade-off achieved using the proposed method. \\
 \begin{table}[h!]
\centering
\small
\begin{tabular}{p{3.5cm}|p{1.4cm}p{1.4cm}p{1.4cm}p{1.4cm}p{1.2cm}p{1.4cm}}
 \hline
 {\bf Ablations} & MOTA$\uparrow$ & MOTP$\uparrow$ & P \%$\uparrow$ & R \%$\uparrow$ & IDs$\downarrow$ &FPS$\uparrow$ \\
%  & \% &\% & \% & \% & & &\\
\hline
%{\bf Ablations}& & & &  & \\
%\hline
Proposed method  &  \textbf{81.2} & 80.0& 93.0 & 91.1& 16 & 13.5\\
\hline 
{\bf Proposals} \\
No track proposals &  80.0 & 79.9 & 92.8 & 90.5& 35 & 13.4\\
\hline 
{\bf Data association}\\
Position only  &  79.4 & 80.0 & 91.9 & 91.9& 46 & \textbf{13.9}\\
Raw ROI-pooled feat. &  80.1 & 80.3 & 93.3 & 89.6& {\bf 12} & 10.5\\
\hline
\end{tabular}
\caption{ KITTI Benchmark ablation study: the proposed method (first row); without tracks proposals (second row); alternative data association: position-based association (third row), position plus appearance based on  raw Faster R-CNN features (fourth row)  }
\label{table:comp}
\end{table}

\noindent  {\em Comparative Analysis:}
To provide a fair comparison with the state of the art, we mainly focus on published methods which have access to comparable data and annotation. Thus, we omit monocular 3D tracking methods and methods which use LiDAR point clouds.

For evaluation, we use KITTI test set and submit results to the evaluation server. %In Table \ref{table:sub_hota} we report Multi-Object Tracking Accuracy (MOTA) \cite{hota}. 
Table \ref{table:sub_hota} provides a general overview on how our proposed method compares to other vision based methods published in the literature.
For all methods not incorporating  detection time in their performance evaluation we added the cost of   Faster R-CNN.
Results show that our proposed method is able to compete with other well performing methods, while, at the same time, achieving near real-time inference  for both detection and tracking, highlighting the advantage of addressing the tasks jointly. Our method, additionally, is fully online, and requires no additional labels.

\begin{table}[h!]
\centering
\small

\begin{tabular}{p{2.0cm}|p{1.7cm}p{1.7cm}p{1.7cm}p{1.1cm}|p{1.1cm}|p{1.1cm}|p{1.0cm}}
 \hline
 {\bf Method} & MOTA\%$\uparrow$ & MOTP\%$\uparrow$ &  MT\%$\uparrow$ & IDs$\downarrow$&3D GT&Online &FPS$\uparrow$ \\
 \hline\hline
Our method  & 81.6   & 80.1  &68.3  &401& &  \checkmark&{\bf 15}\\
\hline

SMAT\cite{smat}& 83.6& 85.9 &62.8&198&&\checkmark&5\\
TuSimple \cite{tusmiple}& 86.3&84.1&71.1&292&&&5\\
QD \cite{quasidense} & 84.9&84.9&69.5&313&&\checkmark&5.8\\
MASS\cite{mass}& 84.6&85.4&74.0&353&&\checkmark&10\\

MOTBP\cite{motbp}& 82.7&85.5&72.6&934&\checkmark& \checkmark& 2.5\\
\hline
\end{tabular}
\caption{ KITTI Tracking benchmark from KITTI evaluation server after submitting the results on the test set for the proposed method}
\label{table:sub_hota}
\end{table}
\subsection{MOT Benchmark} 
\noindent{\em Joint Detection and Tracking}. We use MOT challenge \cite{mot} benchmarks which are the de facto standard in object tracking literature. We focus on MOT17 and MOT20 which offer different levels of difficulty and crowded scenes.
We use the base detector trained on COCOtrain2017 and fine-tune it on MOT17Det, for MOT17 and fine-tune it again on MOT20 training set for MOT20.
Similarly to the above experiment, we attach the embedding head and fine-tune on tracking ground-truth in both cases. In order to account for the crowded scenes, we set the number of proposals to 50 in MOT17 and 100 for MOT20 experiments.

\noindent{\em Comparative Analysis:}
in Table \ref{table:mot17}, to provide a fair comparison, we report our results on the test set alongside results achieved by Tracktor++ and SORT \cite{sort} after accounting for the detection time.
Results of MOT17 benchmark suggest that our proposed method is able to achieve a responsive performance for both tasks jointly while maintaining a speed-accuracy trade-off. While results on MOT20, albeit consistent in terms of comparative analysis, demonstrate a clear degradation in the inference time for all methods, caused by the high resolution video frames and the large number of objects (31 in MOT17 vs 170 in MOT20). Tackling such dense scenes is the aim of our future work efforts.
\begin{table}[h!]
\centering
\small
\begin{tabular}{p{1.8cm}p{1.2cm}p{1cm}p{1.4cm}p{1.4cm}p{1.4cm}p{1.4cm}p{1.4cm}}
 \hline
  {\bf Method} & MOTA$\uparrow$ & IDF1$\uparrow$ & MOTP$\uparrow$ & P \%$\uparrow$ & R \%$\uparrow$  & IDs $\downarrow$ &FPS $\uparrow$ \\ 
 \hline 
\multicolumn{7}{c}{MOT17}\\
\hline
Ours  &  49.4 & 45.1 & 77 & 88.3& 58.1& 5589&  5.37\\
\hline
T.++\cite{nobells} & 53.3 & 52.3 & 78 & 96.3 & 56  & 2072& 1.25 \\
SORT\cite{sort} & 43.1 & 39.8 & 77.8 & 90.7 & 49 & 4852& 7.1\\
\hline
\multicolumn{7}{c}{MOT20}&\\
\hline
Ours & 44.7  &39.1 &76.2&92.5 &  49.5&4171&2.3 \\
\hline
T++\cite{nobells}  & 50.8 & 52.1 & 76.8 & 84.7 & 62.7  & 2751& 0.19 \\
SORT\cite{sort} & 42.7 & 45.1 & 78.5 & 90.2 & 48.8 & 4470& 6.6\\
\hline
\end{tabular}
\caption{ MOT17 and MOT20 Tracking benchmarks obtained from MOT CHALLENGE server after submitting the results on the test set for the proposed method  }
\label{table:mot17}
\end{table}
\subsection{Inference Time Analysis}
In order to better understand the computational cost breakdown across different datasets, we provide a finer-level analysis of the inference time of the joint framework to highlight detection time and tracking time and other parameters which have an influence on the overall time (see Table \ref{table:inf_time}).
While detection time $d\_time$ relies solely on the image dimensions and the number of proposals used, tracking time $t\_time$  is also affected by the average number of objects. Tracking time is dominated by building the distance matrix, and calculating the cosine distance between high-dimensional vectors (the embeddings), while solving the linear assignment using the Hungarian algorithm adds only a marginal cost.
\begin{table}[h!]
\centering
\small
\begin{tabular}{p{1.5cm}p{1.7cm}p{1.5cm}p{1.8cm}p{1.8cm}p{1.6cm}p{1.4cm}}
 \hline
  Dataset & Resolution (average) & \# tracks & \# proposals & d\_time (ms) & t\_time (ms)  &FPS \\ 
 \hline
KITTI &$1242\times375$&7.4&27.4&60.2&5.2&15\\
MOT17 &$1737\times 994$& 22.2& 72.2&94&50&5.3\\
MOT20 &$1394\times907 $& 64.9& 164 & 84 & 315 & 2.3\\
\hline
\end{tabular}
\caption{Average inference time breakdown and other average indicators measured on the test sets of each of the experimented dataset}
\label{table:inf_time}
\end{table}
\section{Conclusion and Future Work}
In this work,  we have proposed a novel and efficient joint object detection and tracking algorithm and discussed the importance of multi-task learning in solving similar visual tasks. Such joint processing is being increasingly adopted in the recent literature with additional inspirations coming from learning using privileged information framework \cite{jointbdd}. New large scale datasets\cite{waymo} provide multi-task annotations to fuel these methods with unprecedented amount of data in the autonomous  driving domain.
%(compared to KITTI size).

Our obtained results demonstrate the benefit of using a simple method with a modular internal structure such as Faster R-CNN in striking a reasonable speed-accuracy trade off, and thus, achieving efficient inference (consistently several-fold faster than other methods) while in the same time delivering competitive accuracy.
Future work will tackle specifically  real-world scenarios, in the autonomous navigation field. This will allow us to fully appreciate the computational benefits of our approach, compared with competing methods. %with possibly tens of different classes of objects in a standard urban scene. %Such trend is already verified in the new datasets which cover a wider range of object classes. 
Additionally,  wider object categories will introduce additional clutter in the scene, for which efficiency aspects need to be pushed further to account for such real-world scenarios.
\bibliographystyle{splncs04}
\bibliography{egbib}

\end{document}